# Capability-Conditioned Scaffolding for Professional Human–LLM Collaboration

**Sen Yang[1] · Yinglei Ma[2]**
[1] University College London [2] Fudan University

## Abstract

Large language model (LLM) personalization typically adapts outputs to user preferences and style but does not account for differences in user evaluation capacity across domains of expertise. This limitation can encourage Professional Domain Drift (PDD), where users rely on AI-generated reasoning in domains they cannot reliably evaluate. We introduce Capability-Conditioned Scaffolding (CCS), a typed framework that partitions expertise into strong, mixed, and weak domains and conditions intervention behavior on structured capability profiles. A pilot evaluation across multiple MMLU subsets and four LLM substrates shows consistent profile-conditioned intervention behavior, including categorical inversion under profile swapping and selective activation in mixed-domain risk zones. These findings suggest that capability-aware scaffolding can support more reliable professional human–AI collaboration beyond stylistic personalization.

## 1. Introduction

Large language models (LLMs) are increasingly used in professional environments to support writing, analysis, decision-making, and advisory work. Existing personalization approaches have primarily focused on adapting outputs to user preferences, interaction history, or stylistic characteristics through prompting, retrieval augmentation, and alignment-based tuning (Lewis et al., 2020; Ouyang et al., 2022; Park et al., 2023). While these approaches can improve usability and interaction fluency, they generally do not account for differences in user evaluation capacity across domains of expertise.

This limitation becomes particularly consequential in professional settings, where LLMs can encourage users to operate beyond the boundaries of their substantive expertise. Prior studies have shown that AI systems can improve performance within familiar domains while simultaneously increasing failure rates near the boundary of user expertise, particularly in complex professional tasks where users struggle to reliably validate outputs (Dell'Acqua et al., 2023; Vaccaro et al., 2024). We refer to this phenomenon as Professional Domain Drift (PDD), describing situations in which AI-assisted judgment extends into adjacent domains that users cannot adequately evaluate despite possessing broader professional authority or contextual familiarity.

The central challenge of PDD is not simply incorrect output generation, but the erosion of reliable evaluative oversight during human–AI collaboration. Research on automation reliance and algorithm-in-the-loop decision-making has repeatedly shown that human oversight can become procedural rather than substantive, especially when systems produce fluent or authoritative outputs (Lee & See, 2004; Green & Chen, 2019; Green, 2022). Existing interventions have therefore introduced user-side reflection mechanisms, including cognitive forcing functions intended to encourage slower or more deliberate verification behavior (Buçinca et al., 2021). However, these approaches implicitly assume that users recognize when they are operating outside their evaluative competence. In professional environments, this assumption often fails because adjacent-domain tasks may still appear cognitively accessible or professionally legitimate to the user.

Addressing this problem requires moving beyond stylistic personalization toward capability-aware interaction governance. In this paper, we introduce Capability-Conditioned Scaffolding (CCS), a typed architectural framework that conditions LLM intervention behavior on structured representations of user capability. CCS partitions expertise into strong, mixed, and weak domains and adjusts intervention intensity through a deterministic decision matrix. Rather than treating the user as a generalized professional persona, the framework models expertise boundaries as operational constraints that shape how the system responds to different categories of requests.

The proposed framework incorporates consistency guardrails intended to detect mismatches between declared expertise and prompt-level evidence. These mechanisms are designed to regulate intervention behavior in mixed-domain contexts where users possess partial familiarity but limited evaluative reliability. Unlike oversight architectures that rely on separate evaluator systems (Bowman et al., 2022; Burns et al., 2023), CCS embeds capability-conditioned scaffolding directly within the response-generation process through structured prompting and routing policies.

To evaluate the feasibility of this approach, we conduct a pilot study across multiple MMLU subsets and four LLM substrates under controlled single-turn conditions. The results show consistent profile-conditioned intervention behavior across models, including categorical inversion under profile swapping and selective activation within mixed-domain risk zones. While the present study is limited to prompt-level architectural validation, the findings provide initial evidence that typed capability scaffolding can systematically influence human–LLM interaction behavior beyond conventional stylistic personalization.

This paper makes three contributions. First, it introduces Professional Domain Drift as a capability-boundary problem in professional human–LLM interaction. Second, it proposes Capability-Conditioned Scaffolding as a typed architectural framework for capability-aware intervention. Third, it provides pilot empirical evidence demonstrating that intervention behavior can be systematically modulated through structured capability profiles across multiple LLM substrates.

## 2. Related Work

### 2.1 Personalization Without Capability Awareness

Personalization in large language models (LLMs) has primarily focused on adapting outputs to user preferences, interaction history, or stylistic characteristics. Existing approaches include retrieval-augmented generation (Lewis et al., 2020), instruction alignment through reinforcement learning from human feedback (Ouyang et al., 2022), and persona-oriented behavioral prompting (Park et al., 2023). These methods improve interaction fluency and contextual relevance but generally treat the user as a preference-bearing or stylistic entity rather than as an actor with uneven evaluative capacity across domains.

As personalization systems become increasingly integrated into professional workflows, this limitation becomes more consequential. Existing personalization pipelines can adapt tone, format, and contextual memory, yet they typically lack explicit representations of where users can or cannot reliably evaluate AI-generated reasoning. As a result, personalization may inadvertently reinforce overreliance in adjacent domains by increasing interaction fluency without calibrating intervention intensity to capability boundaries.

CCS extends this literature by introducing typed capability partitions as an explicit architectural component. Rather than modeling personalization solely as preference alignment or behavioral imitation, the framework conditions system behavior on structured representations of user expertise across strong, mixed, and weak domains.

### 2.2 Oversight and Reliance in Human–AI Collaboration

Research on human–AI collaboration has consistently shown that effective oversight is difficult to maintain when users interact with fluent and authoritative AI systems. Foundational work on trust in automation demonstrated that excessive reliance could emerge when users miscalibrate system competence or fail to sustain active verification behavior (Lee & See, 2004). Subsequent studies on algorithm-in-the-loop decision-making further showed that human oversight often becomes procedural rather than substantive in practice (Green & Chen, 2019; Green, 2022).

Recent work has explored intervention mechanisms intended to mitigate overreliance. Cognitive forcing functions, for example, introduce deliberate friction to encourage users to engage more critically with AI-generated outputs (Buçinca et al., 2021). Studies on human–AI complementarity similarly suggest that collaborative performance depends on calibrated trust and domain-sensitive delegation rather than unconditional reliance (Vaccaro et al., 2024). At the same time, empirical evidence indicates that AI systems may improve performance within familiar domains while increasing failure rates near the boundary of user expertise (Dell'Acqua et al., 2023).

However, many existing interventions rely implicitly on user self-awareness. They assume that users recognize when they are operating outside their evaluative competence and will respond appropriately to reflective prompts or verification cues. In professional environments, this assumption may not hold because adjacent-domain tasks often remain cognitively accessible and institutionally legitimate to the user. CCS addresses this gap by shifting intervention responsibility from user-initiated reflection toward capability-conditioned system behavior.

### 2.3 Capability-Aware Intervention and Embedded Scaffolding

Related work on scalable oversight and weak-to-strong supervision has examined how weaker evaluators can supervise more capable AI systems through structured decomposition or external evaluation architectures (Bowman et al., 2022; Burns et al., 2023). Other approaches, such as Constitutional AI, introduce principle-based behavioral constraints during generation (Bai et al., 2022). These frameworks primarily conceptualize oversight as a separable evaluative layer operating alongside or above the generation process.

CCS differs from these approaches in two respects. First, it focuses specifically on capability-boundary regulation in professional human–AI interaction rather than on generalized alignment or scalable supervision. Second, the framework embeds intervention logic directly within the response-generation process through typed capability profiles and routing policies rather than through external evaluator architectures.

The framework also draws conceptually from research on professional jurisdiction and sociotechnical interaction, which emphasizes that professional authority and evaluative competence do not always align uniformly across adjacent domains (Abbott, 1988). Rather than treating expertise as globally transferable, CCS models capability as domain-partitioned and operationally bounded. This framing allows intervention behavior to vary according to the relationship between prompt content and declared expertise rather than according to a single global user profile.

### 2.4 Task Regimes and the Scope of Capability-Aware Intervention

Empirical evidence on human–AI joint performance divides sharply by task regime. In simple-task contexts such as factual recall or binary classification, recent meta-analytic work shows that human–AI combinations often perform worse than the best of humans or AI alone, with synergy losses concentrating in decision-making tasks where one party substantially outperforms the other (Vaccaro et al., 2024). Empirical evidence in complex-task contexts such as differential diagnosis and structured advisory reasoning points in more mixed directions: hybrid configurations have shown advantages over individual experts or standalone systems in some clinical settings, while comparative validation studies in evidence synthesis tasks have found that contemporary LLMs remain unreliable for autonomous use and require human oversight.

This task-regime distinction matters for the scope of capability-aware intervention. Simple-task regimes where LLM-alone performance is reliable are not the deployment context in which CCS-style scaffolding has substantive room to operate. The architectural claim targets complex-task professional contexts where joint-output non-additivity is empirically established and where users may possess partial familiarity without reliable evaluative capacity. The present pilot evaluation uses MMLU subsets as a controlled proxy for this regime, with the understanding that benchmark prompts cannot fully reproduce the contextual complexity of real professional advisory environments.

Taken together, prior work has examined personalization, oversight, and human–AI reliance from multiple perspectives, but capability-aware intervention remains comparatively under-specified. CCS contributes to this emerging area by proposing a typed architectural framework that conditions intervention behavior on structured representations of user evaluation capacity.

## 3. Core CCS Mechanisms

CCS is designed to regulate human–LLM interaction behavior according to structured representations of user evaluation capacity. Rather than treating personalization as stylistic adaptation alone, CCS models expertise boundaries as operational constraints that shape intervention behavior across different categories of requests. The framework focuses particularly on situations where users possess partial familiarity with a domain but lack sufficient evaluative reliability to independently validate model outputs.

The current study evaluates the prompt-level architectural core of CCS under controlled single-turn conditions. Accordingly, this section focuses on the mechanisms directly examined in the pilot evaluation, namely typed capability partitioning, intervention routing, and consistency-based activation.

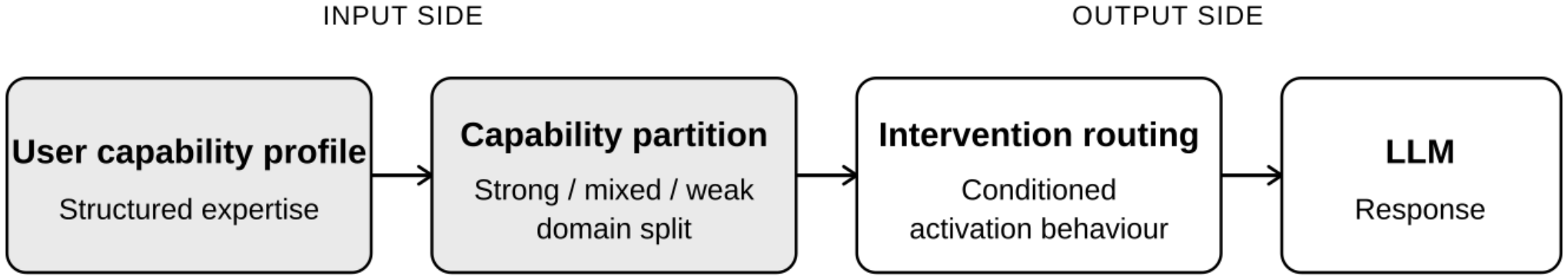

***Figure 1. CCS architecture overview.*** *Intervention behavior is conditioned on structured representations of user evaluation capacity through a four-stage pipeline. The input side specifies user capability and partitions expertise into strong, mixed, and weak domains. The output side routes intervention intensity and generates the LLM response accordingly.*

### 3.1 Typed Capability Profiles

CCS represents user expertise through a domain-partitioned capability profile. Instead of assuming globally transferable competence, the framework models expertise as unevenly distributed across domains. Capability profiles are partitioned into three operational categories.

**Strong domains** refer to areas in which the user possesses reliable evaluative capacity grounded in substantive expertise. In these contexts, intervention behavior is minimized in order to preserve interaction efficiency and reduce unnecessary friction. **Mixed domains** represent adjacent or partially familiar areas where users possess contextual understanding but may lack the ability to reliably evaluate AI-generated reasoning. These domains constitute the **operational PDD-risk zone** of the framework, since users may overestimate their evaluative competence while still perceiving the task as professionally accessible. **Weak domains** refer to areas outside the user's substantive expertise, where the user functions primarily as a lay evaluator. In these contexts, the framework increases intervention intensity and transparency signalling.

This partitioned representation allows intervention behavior to vary according to the relationship between prompt content and declared expertise rather than according to a single generalized user identity.

### 3.2 Capability-Conditioned Intervention Routing

CCS adjusts intervention behavior through a capability-conditioned routing mechanism. The framework evaluates the relationship between prompt content, domain classification, and inferred reasoning complexity in order to determine the appropriate level of scaffolding. In strong-domain contexts, the system is designed to remain minimally intrusive unless prompt ambiguity or unusually high reasoning uncertainty is detected. In weak-domain contexts, intervention behavior becomes substantially more active, including increased uncertainty disclosure and capability-boundary signaling. Mixed domains occupy an intermediate position in which intervention intensity depends more heavily on prompt-level evidence and conceptual structure. The routing mechanism therefore functions as a structured alternative to purely user-initiated verification. Rather than assuming that users will recognize when additional caution is required, CCS conditions intervention behavior directly on capability partitions and domain-sensitive risk assessment.

### 3.3 Consistency-Based Guardrails

A central mechanism within CCS is the use of consistency-based guardrails that compare declared expertise against prompt-level evidence. The framework is designed to detect situations in which the conceptual structure of a request diverges from the user's stated capability profile. This mechanism is particularly important within mixed-domain contexts. Adjacent-domain prompts often contain terminology or reasoning structures that appear superficially familiar while nevertheless requiring evaluative capacities outside the user's reliable expertise. CCS therefore evaluates whether prompt characteristics align with the declared capability partition before determining intervention intensity.

Importantly, the framework does not treat capability declarations as absolute or globally transferable. Instead, intervention behavior remains conditional on the relationship between profile structure and prompt evidence. This allows the system to distinguish between adjacent domains that share genuine conceptual overlap and those that merely appear contextually related from the user's perspective.

### 3.4 Extended Multi-Turn Components

Beyond the single-turn mechanisms evaluated in the present study, CCS also includes proposed extensions for multi-turn interaction settings. These extensions include longitudinal calibration mechanisms intended to track delegation patterns, interaction continuity, and accumulated evidence of evaluative reliability across sessions. Such components are intended to support dynamic adjustment of intervention intensity over time, particularly in repeated professional workflows where trust calibration and delegation behavior evolve across interactions. However, these mechanisms are not evaluated in the current pilot study and remain part of the broader architectural roadmap for future work.

## 4. Pilot Evaluation

### 4.1 Evaluation Objective

The present evaluation examines whether CCS produces systematic profile-conditioned intervention behavior under controlled prompt-level conditions. The study focuses specifically on the core mechanisms introduced in Section 3, namely typed capability partitioning, intervention routing, and consistency-based activation. The evaluation does not measure downstream decision quality, user trust calibration, or real-world professional outcomes. Instead, the objective is to determine whether structured capability profiles influence intervention behavior consistently across domains and model substrates.

### 4.2 Evaluation Conditions

We evaluated CCS using 480 question-condition pairs derived from multiple subsets of the Massive Multitask Language Understanding (MMLU) benchmark. The evaluation compared three primary conditions. The first condition consisted of a baseline setting without capability-conditioned scaffolding. The second condition used **PCS-NLP**, an NLP-oriented capability profile in which technical and computational domains were classified as strong expertise areas. The third condition used **PCS-LitProf**, a profile-inverted literature-oriented configuration in which humanities-related domains were treated as strong expertise areas while technical domains were treated as weak domains. All evaluations were conducted under controlled single-turn conditions using deterministic generation settings. Full capability profile specifications are provided in Appendix A.

**Table 1. Core CCS mechanisms evaluated in the present pilot study.**

| Mechanism | Functional Role | Evaluated in Pilot |
|---|---|---|
| Typed capability profiles | Partition expertise into strong, mixed, and weak domains | Yes |
| Intervention routing | Adjust scaffolding intensity according to capability boundaries | Yes |
| Consistency-based guardrails | Compare prompt evidence against declared expertise | Yes |
| Multi-turn calibration extensions | Track longitudinal interaction and delegation patterns | No |

### 4.3 Capability Partition Design

The selected MMLU subsets were partitioned into strong, mixed, and weak domains relative to the NLP-oriented profile.

Machine Learning and College Computer Science were categorized as strong domains. Professional Psychology, Formal Logic, and Econometrics were categorized as mixed domains representing potential PDD risk zones. Clinical Knowledge, Medical Genetics, Professional Law, Philosophy, and European History were categorized as weak domains. The mixed-domain configuration was included to evaluate whether intervention behavior varied within formally identical capability partitions according to prompt-level conceptual alignment. Partition rationale and subset selection criteria are detailed in Appendix B.

### 4.4 Cross-Substrate Replication

To examine substrate portability, the evaluation was replicated across four LLM substrates representing multiple vendors and model generations, including Claude Sonnet 4.5, Claude Haiku 4.5, GPT-4.1, and gpt-5.5. The evaluation focused on directional consistency of intervention behavior across strong, mixed, and weak capability partitions rather than on benchmark-style performance optimization.

### 4.5 Evaluation Boundaries

Several limitations constrain the scope of the present evaluation. First, the study is restricted to controlled single-turn interactions and therefore does not evaluate longitudinal delegation behavior or adaptive trust calibration across repeated sessions. Second, the evaluation measures intervention activation behavior rather than human behavioral outcomes or professional decision quality. Third, benchmark-style prompts cannot fully reproduce the contextual complexity of real professional environments.

Accordingly, the present study should be interpreted as a pilot evaluation of prompt-level architectural behavior rather than a complete validation of capability-aware governance in deployed professional systems.

## 5. Results

### 5.1 Profile-Conditioned Intervention Behavior

The primary evaluation examined whether intervention behavior changed systematically when identical prompts were paired with different capability profiles. Under the PCS-NLP profile, intervention activation within strong ML/CS subsets remained low at 3.3% (1/30), while weak-domain subsets produced substantially higher activation rates, including 100% activation in medical and legal subsets (30/30) and 90.0% activation in humanities subsets (27/30).

**Table 2. Profile inversion results across domain partitions.**

| Condition | ML / CS (Strong) | Medical / Legal (Weak) | Humanities (Weak) |
|---|---|---|---|
| PCS-NLP | 3.3% (1/30) | 100% (30/30) | 90.0% (27/30) |
| PCS-LitProf | 100% (30/30) | 73.3% (22/30) | 0% (0/30) |

When the capability profile was inverted toward the PCS-LitProf configuration, the activation pattern correspondingly shifted. Intervention activation in ML/CS subsets increased from 3.3% (1/30) to 100% (30/30), while humanities activation decreased from 90.0% (27/30) to 0% (0/30). The prompt sets remained identical across conditions, with only the capability profile modified.

Fisher's exact tests showed strong separation between the two profile conditions for both ML/CS and humanities subsets ($p < 10^{-13}$ for both comparisons). Permutation testing on matched prompt pairs further confirmed that the observed inversion pattern was unlikely to emerge under random profile assignment ($p < 10^{-4}$). Overall, intervention behavior remained consistently aligned with the declared capability partition across matched prompt conditions.

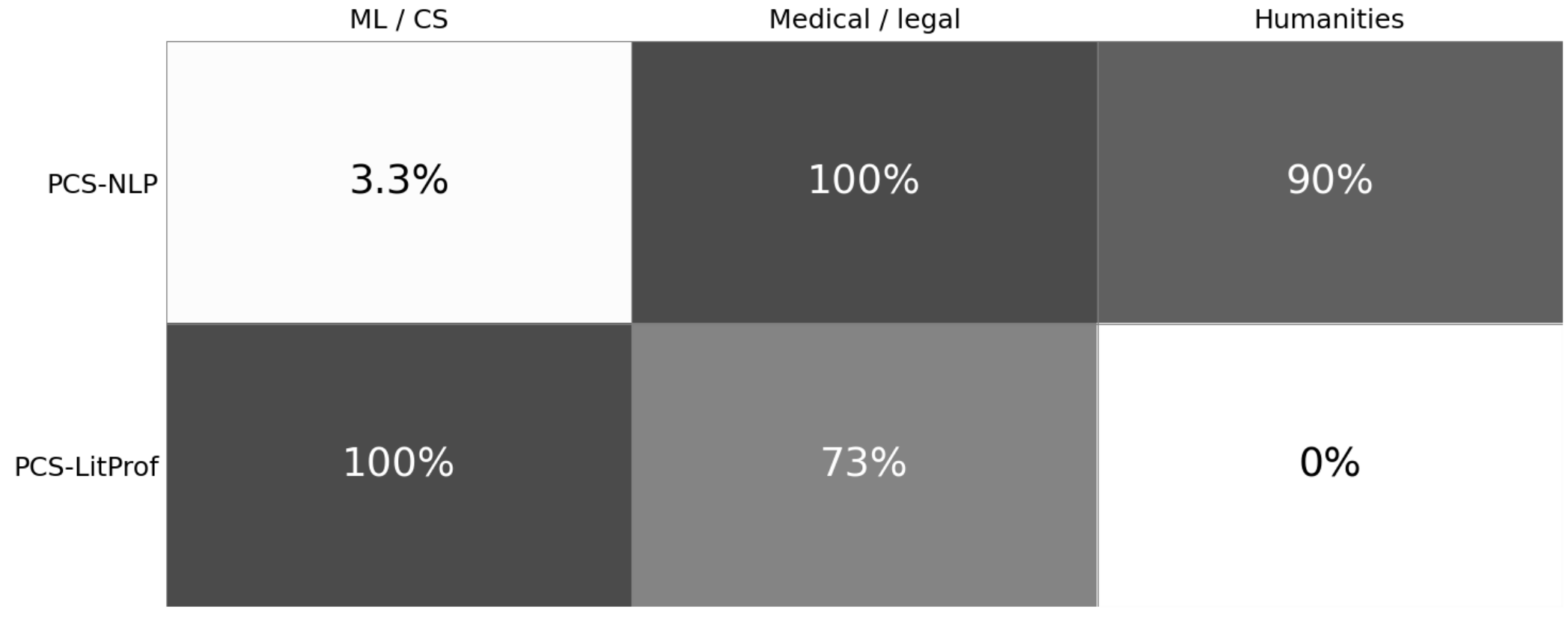

***Figure 2. Profile-conditioned intervention activation across NLP-oriented (PCS-NLP) and literature-oriented (PCS-LitProf) capability profiles.*** *Activation rates invert categorically when only the profile is modified; prompts remain identical across conditions. Darker cells indicate higher activation rate.*

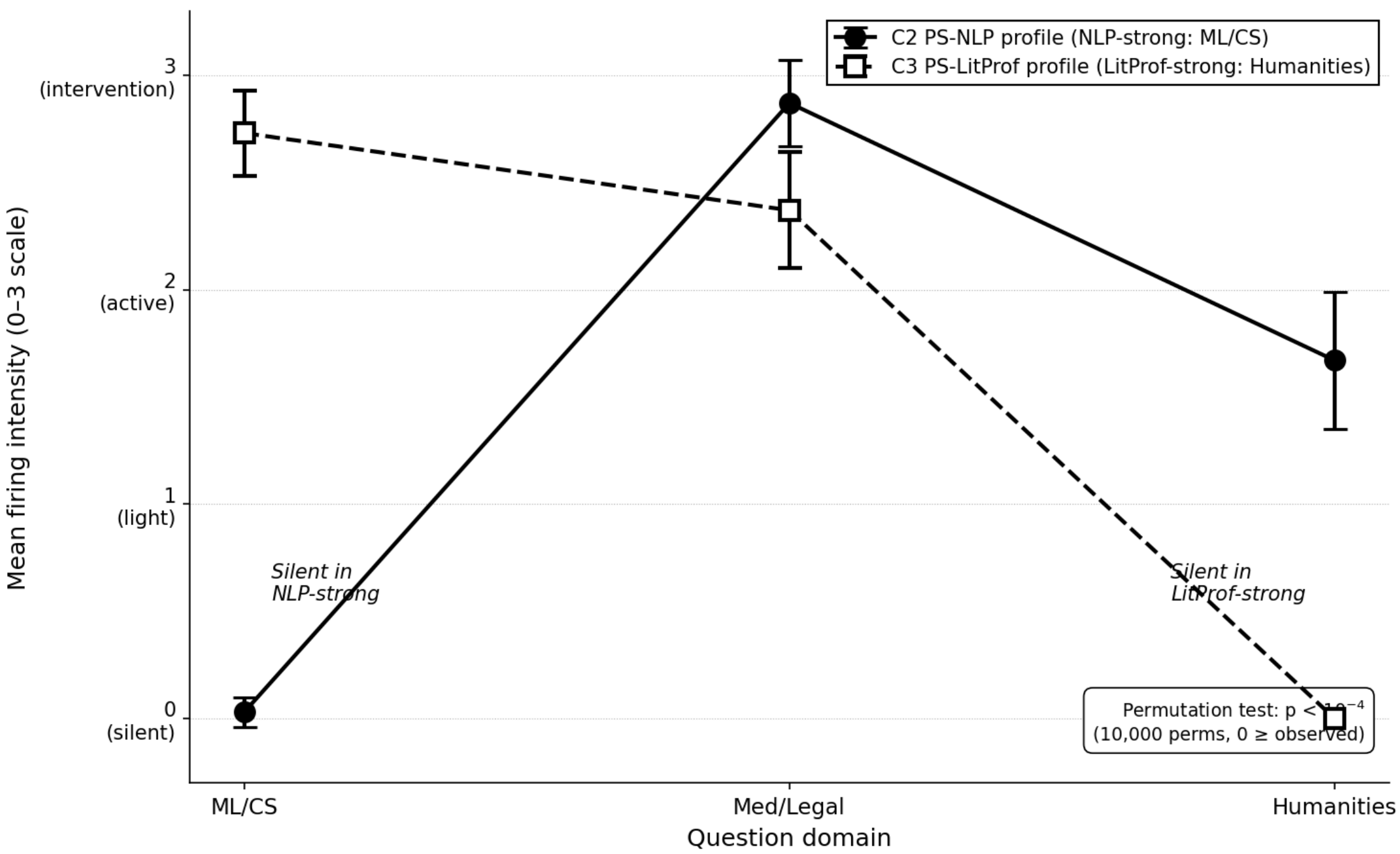

***Figure 3. Mean intervention firing intensity partitions under profile conditions.*** *Error bars represent 95% confidence intervals. Permutation test on matched prompt pairs returned p < 10^-4 (10,000 permutations; 0 ≥ observed). Annotations indicate the "silent" condition (firing intensity = 0) in each profile's strong domain.*

### 5.2 Mixed-Domain Activation Patterns

The mixed-domain evaluation examined whether intervention behavior differed within formally identical capability partitions. Under the PCS-NLP profile, Professional Psychology, Formal Logic, and Econometrics were all categorized as mixed domains. However, intervention activation rates varied substantially across these subsets.

Professional Psychology produced an activation rate of 73% (22/30), whereas Formal Logic produced 13% activation (4/30) and Econometrics produced 3% activation (1/30). In the baseline condition without capability-conditioned scaffolding, no intervention activation was observed across mixed-domain subsets.

**Table 3. Within-partition activation divergence.**

| Mixed-Domain Subset | Intervention Activation |
|---|---|
| Professional Psychology | 73% (22/30) |
| Formal Logic | 13% (4/30) |
| Econometrics | 3% (1/30) |

Although all three subsets shared the same formal capability partition, intervention behavior differed substantially across prompt categories. Formal Logic and Econometrics frequently remained minimally intrusive despite belonging to the same declared mixed-domain category as Professional Psychology.

These results indicate that intervention behavior was sensitive not only to partition labels but also to prompt-level conceptual alignment with the declared capability profile.

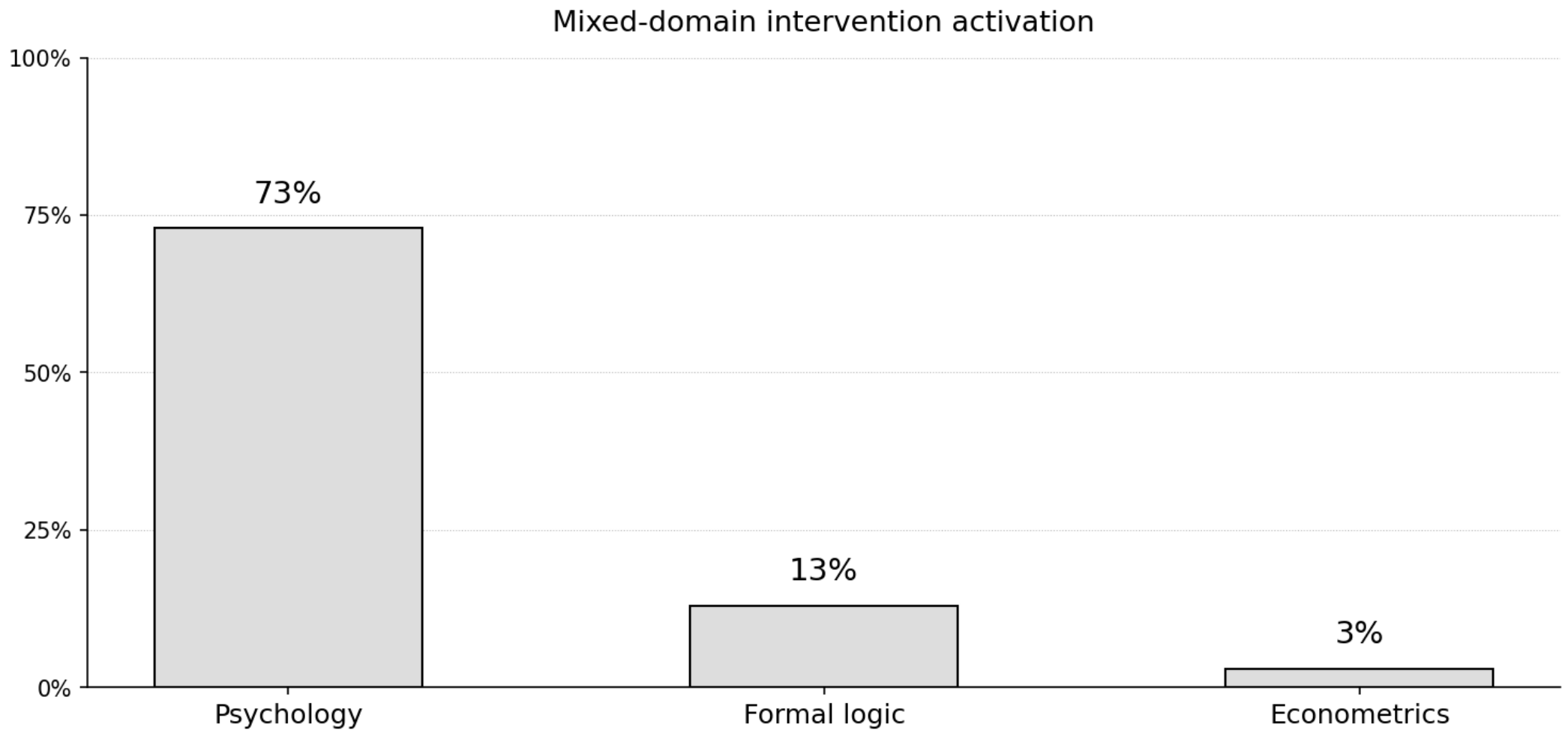

***Figure 4. Intervention activation rates within the mixed-domain partition.*** *Despite identical formal partition labels, activation differs substantially across Professional Psychology, Formal Logic, and Econometrics, indicating sensitivity to prompt-level conceptual alignment beyond static domain categories.*

**5.3 Cross-Substrate Consistency**

Across all substrates, intervention activation remained consistently low within strong domains and substantially higher within weak domains. For strong ML/CS subsets, activation rates ranged from 0% to 3.3% across models. For weak medical and legal subsets, activation rates ranged from 93.3% to 100%. Humanities activation ranged from 76.7% to 100% depending on substrate configuration.

**Table 4. Cross-substrate activation rates across four LLM models.**

| Domain Partition | Sonnet 4.5 | Haiku 4.5 | GPT-4.1 | gpt-5.5 |
|---|---|---|---|---|
| ML / CS (Strong) | 3.3% | 0% | 0% | 0% |
| Medical / Legal (Weak) | 100% | 93.3% | 100% | 100% |
| Humanities (Weak) | 90.0% | 76.7% | 96.7% | 100% |

Despite moderate variation in activation intensity and formatting behavior, the overall routing pattern remained directionally stable across substrates.

**5.4 Borderline Cases and Annotation Consistency**

Within the PCS-NLP condition, only 1 of 30 ML/CS questions produced intervention activation. Inspection of this case showed that the prompt involved ambiguity in Bayesian network parameter counting, suggesting that activation was associated with uncertainty in prompt interpretation rather than broad strong-domain escalation.

Under the PCS-LitProf condition, all 30 ML/CS questions produced intervention activation, consistent with the profile inversion design in which ML/CS domains were categorized as weak expertise areas.

Annotation consistency remained high across all capability-conditioned conditions. Structured intervention formatting was maintained across nearly all outputs, with only minor formatting variation observed in selected Sonnet 4.5 medical-domain responses.

## 6. Discussion

### 6.1 Interpretation of Findings

The present study examined whether capability-conditioned scaffolding can systematically modulate intervention behavior according to structured representations of user expertise. Across all evaluation conditions, the results showed consistent profile-conditioned routing behavior under controlled single-turn settings. Identical prompts produced substantially different intervention patterns when paired with different capability profiles, while mixed-domain evaluations showed differentiated activation behavior within formally identical partition labels.

These findings are consistent with the central architectural premise of CCS. Rather than treating personalization as stylistic adaptation alone, the framework conditions intervention behavior on capability boundaries and prompt-level conceptual alignment. The mixed-domain results are particularly important in this regard. Intervention behavior varied substantially across Professional Psychology, Formal Logic, and Econometrics despite all three subsets being assigned to the same formal partition category. This pattern suggests that the framework responds not only to static domain labels but also to relationships between prompt structure and declared expertise.

The results also indicate that capability-conditioned intervention can be implemented at the prompt-interaction level across multiple LLM substrates. Activation intensity and surface formatting varied moderately across vendors and model generations, but the overall routing pattern—low activation in strong domains, high activation in weak domains, and content-sensitive activation in mixed domains—remained directionally stable across all four tested systems. This stability suggests that CCS operates at an architectural layer above vendor-specific prompt conditioning rather than as a model-family-specific behavior. Cross-substrate portability is a preliminary finding at the present sample, and broader vendor coverage is required to establish the boundaries of this generalization.

### 6.2 Implications for Human–AI Collaboration

Current personalization approaches primarily optimize interaction fluency, stylistic adaptation, and contextual continuity. The findings presented here suggest that professional human–AI collaboration may also require mechanisms that account for uneven evaluative capacity across domains.

This issue becomes particularly relevant in adjacent-domain contexts associated with PDD. In such settings, users may possess sufficient contextual familiarity to engage confidently with AI-generated outputs while lacking the expertise necessary to independently evaluate underlying reasoning. Existing intervention approaches often rely on user self-awareness and reflective verification behavior. CCS instead explores a system-side alternative in which intervention behavior is conditioned directly on structured capability partitions.

More broadly, the results suggest that personalization may benefit from moving beyond preference alignment alone toward forms of capability-aware interaction governance. Under this framing, personalization is not only a matter of adapting outputs to how users prefer to interact, but also of calibrating intervention behavior according to where users can reliably evaluate model-generated reasoning.

### 6.3 Architectural Positioning

CCS should not be interpreted as a replacement for external oversight systems, factual verification pipelines, or alignment-oriented safety methods. The framework does not determine whether generated outputs are objectively correct, nor does it attempt to solve generalized truthfulness or reasoning reliability problems.

Instead, CCS operates as an interaction-layer architecture designed to regulate how intervention behavior changes across capability boundaries. The framework focuses specifically on situations in which users interact with AI systems outside their strongest evaluative domains. Under this interpretation, capability-conditioned scaffolding functions as a governance-oriented interaction mechanism rather than as a standalone safety or verification system.

The present study also evaluates only the prompt-level architectural floor of the framework. The proposed multi-turn extensions, including longitudinal calibration and behavioral evidence accumulation, remain unevaluated in the current work and should therefore be treated as future architectural directions rather than validated system components.

### 6.4 Limitations and Future Work

Several limitations constrain the interpretation of the present findings. First, the evaluation was conducted entirely under controlled single-turn conditions and therefore cannot assess longitudinal trust calibration, adaptive delegation behavior, or repeated professional workflows. Second, the study measures intervention activation behavior rather than downstream human behavioral outcomes or decision quality. Third, the evaluation relies primarily on benchmark-style prompts and synthetic capability profiles rather than ecological deployment settings involving real professionals.

Future work should therefore extend CCS into multi-turn interaction environments and evaluate whether capability-conditioned intervention influences user reliance, verification behavior, or collaborative decision quality over time. Additional work is also needed to examine ecological deployment scenarios across professional domains such as peer review, legal reasoning, medical consultation, and advisory workflows.

To support deployment-context evaluation, we have developed and openly released the PDD-Scenarios corpus, a synthetic persona corpus comprising three professional role archetypes (management consultant, corporate lawyer, and NLP/ML researcher), each spanning strong, PDD-risk-zone, and clearly outside content zones. The corpus is intended as a community-extensible testbed for evaluating capability-aware intervention architectures beyond benchmark-style prompts.

More generally, the present study should be interpreted as a pilot investigation into capability-aware interaction governance rather than as a complete validation of professional AI oversight. The findings nevertheless suggest that structured capability profiles can serve as a viable architectural mechanism for regulating intervention behavior in professional human–LLM collaboration.

## 7. Conclusion

Large language models are increasingly integrated into professional workflows that extend beyond information retrieval and text generation into analysis, judgment, and advisory support. Existing personalization approaches have substantially improved interaction fluency and contextual adaptation, but they generally do not account for differences in user evaluation capacity across domains of expertise. This limitation becomes particularly consequential in settings where users interact with AI-generated reasoning outside their strongest evaluative domains.

This paper introduced CCS, a typed architectural framework that conditions intervention behavior on structured capability profiles rather than stylistic personalization alone. A pilot evaluation across multiple MMLU subsets and four LLM substrates showed consistent profile-conditioned intervention behavior under controlled single-turn conditions, including categorical inversion under profile swapping, differentiated activation within mixed-domain contexts, and stable routing behavior across multiple model families. While the present study does not constitute deployment-level validation, the findings provide initial evidence that capability-aware intervention can be implemented as a prompt-level interaction mechanism across different LLM substrates. More broadly, this work suggests that personalization in professional human–AI collaboration may require mechanisms that address capability boundaries in addition to stylistic adaptation and preference alignment.

## Appendix A: Capability Profile Specifications

This appendix documents the structure of the two capability profiles used in the pilot evaluation. Profiles are operationalized as structured prompt-level metadata that the CCS framework uses to condition intervention routing.

### A.1 PCS-NLP profile

The NLP-oriented capability profile assigns user expertise across three operational categories:

**Strong domains**: Machine Learning, College Computer Science. The user is treated as possessing reliable evaluative capacity for technical and computational reasoning. Intervention behavior is minimized in these areas.

**Mixed domains** (operational PDD-risk zone): Professional Psychology, Formal Logic, Econometrics. The user is treated as possessing partial familiarity but limited evaluative reliability. Intervention behavior depends on prompt-level conceptual alignment with the user's declared technical background.

**Weak domains**: Clinical Knowledge, Medical Genetics, Professional Law, Philosophy, European History. The user is treated as a lay evaluator. Intervention behavior is escalated, with explicit uncertainty disclosure and capability-boundary signaling.

### A.2 PCS-LitProf profile

The literature-oriented capability profile inverts the technical/humanistic axis:

**Strong domains**: Philosophy, European History (literature-oriented humanities expertise).

**Mixed domains** (operational PDD-risk zone): Professional Psychology, Formal Logic.

**Weak domains**: Machine Learning, College Computer Science, Clinical Knowledge, Medical Genetics, Professional Law.

The PCS-LitProf condition was constructed specifically to test profile inversion: the same prompts that elicit minimal intervention under PCS-NLP should elicit maximal intervention under PCS-LitProf if CCS conditions on the profile rather than on prompt content alone.

### A.3 Intervention routing prompt template

The CCS routing layer receives the capability profile as structured input and conditions LLM behavior through the following template structure:

User capability profile: { strong: [...], mixed: [...], weak: [...] }

Routing policy: For prompts in strong domains, respond directly and minimize intervention markers. For prompts in mixed domains, evaluate prompt-evidence alignment before determining intervention intensity. For prompts in weak domains, include explicit uncertainty disclosure and capability-boundary signaling.

Full prompt scaffolding, intervention markers, and decision-matrix encoding are available from the corresponding author on request.

## Appendix B: MMLU Subset Selection and Partition Rationale

This appendix specifies the criteria used to select MMLU subsets and assign them to capability partitions in the pilot evaluation.

### B.1 Subset selection criteria

MMLU subsets were selected to satisfy three criteria. First, subsets needed to be operationally relevant to professional advisory work, allowing the evaluation to approximate professional reasoning contexts within benchmark constraints. Second, subsets needed to span the strong/mixed/weak partition structure relative to a coherent professional profile (NLP-oriented). Third, subsets needed to include adjacent-domain content that approximates the operational PDD-risk zone described in §3.1.

Ten MMLU subsets were selected: Machine Learning, College Computer Science, Professional Psychology, Formal Logic, Econometrics, Clinical Knowledge, Medical Genetics, Professional Law, Philosophy, and European History.

### B.2 Partition assignment rationale

Partition assignments under the PCS-NLP profile follow the technical/applied distinction characteristic of NLP and computer science training.

**Strong-domain assignments** (ML, College CS): These domains constitute core technical content of NLP and computer science research, where a user with this profile would possess reliable evaluative capacity through substantive training and ongoing engagement.

**Mixed-domain assignments** (Professional Psychology, Formal Logic, Econometrics): These domains share conceptual or methodological proximity to NLP research but require evaluative capacity that does not transfer reliably from technical training alone. Professional Psychology requires clinical and methodological judgment beyond research-method familiarity. Formal Logic shares notational overlap with ML but applies different inference structures. Econometrics shares quantitative methodology but requires domain-specific causal reasoning. These three subsets were selected specifically to enable within-partition divergence testing (Section 5.2).

**Weak-domain assignments** (Clinical Knowledge, Medical Genetics, Professional Law, Philosophy, European History): These domains lie outside the typical training and ongoing exposure of an NLP-oriented user. The user functions primarily as a lay evaluator in these areas.

### B.3 Sample size

Each MMLU subset contributed 30 question items to the evaluation. Across 10 subsets and 3 conditions (baseline, PCS-NLP, PCS-LitProf), this produced 480 question-condition pairs after profile-applicable filtering. Sample sizes of 30 per cell are consistent with pilot-tier statistical detection for categorical effects but are insufficient for fine-grained subset-level effect estimation. Statistical upgrade to $n>100$ per cell is specified for Phase 2 work.